\documentclass[10pt,twocolumn,letterpaper]{article}

\usepackage{cvpr}
\usepackage{times}
\usepackage{epsfig}
\usepackage{graphicx}
\usepackage{amsmath}
\usepackage{amssymb}
\usepackage{url}
\usepackage{bm}

\usepackage{color}
\usepackage{algorithm}
\usepackage{algpseudocode}

\usepackage{enumitem}
\usepackage{mathtools}
\usepackage{float}
\usepackage{multirow}
\usepackage{enumitem}
\usepackage{capt-of}
\usepackage{array,multirow}
\usepackage{physics}
\usepackage{comment}
\usepackage{caption,subcaption}
\usepackage{authblk}
\usepackage{lipsum}
\usepackage{algorithm}
\usepackage{algpseudocode}

\newcommand*{\affaddr}[1]{#1} 
\newcommand*{\affmark}[1][*]{\textsuperscript{#1}}

\usepackage[pagebackref=true,breaklinks=true,letterpaper=true,colorlinks,bookmarks=false]{hyperref}

\cvprfinalcopy 

\def\cvprPaperID{2161} 

\ifcvprfinal\fi
\begin{document}

\setlength{\textfloatsep}{2pt}%
\title{\textbf{\textit{Viewpoint}}-aware Video Summarization}

\author{%
 Atsushi Kanehira\affmark[1], Luc Van Gool\affmark[3,4], Yoshitaka Ushiku\affmark[1], and Tatsuya Harada\affmark[1,2]\\
\affaddr{\affmark[1]The University of Tokyo}, \affaddr{\affmark[2]RIKEN}, \affaddr{\affmark[3]ETH Z\"urich},  \affaddr{\affmark[4]KU Leuven}
 }

\maketitle
\begin{abstract}
This paper introduces a novel variant of video summarization, namely building a summary that depends on the particular aspect of a video the viewer focuses on. We refer to this as {\it viewpoint}. To infer what the desired viewpoint may be, we assume that several other videos are available, especially groups of videos, e.g., as folders on a person's phone or laptop. The semantic similarity between videos in a group vs. the dissimilarity between groups is used to produce viewpoint-specific summaries. For considering {\it similarity} as well as avoiding redundancy, output summary should be (A) diverse,
(B) representative of videos in the same group, and 
(C) discriminative against videos in the different groups.
To satisfy these requirements (A)-(C) simultaneously, we proposed a novel video summarization method from multiple groups of videos. Inspired by Fisher's discriminant criteria, it selects summary by optimizing the combination of three terms (a) inner-summary, (b) inner-group, and (c) between-group variances defined on the feature representation of summary, which can simply represent (A)-(C). 
Moreover, we developed a novel dataset to investigate how well the generated summary reflects the underlying {\it viewpoint}.
Quantitative and qualitative experiments conducted on the dataset demonstrate the effectiveness of proposed method.
\end{abstract}

\vspace{-0.2cm}
\section{Introduction}\label{sec:introduction}

\begin{figure}[t]
\begin{center}
\begin{tabular}{c}
\includegraphics[clip, width=\linewidth]{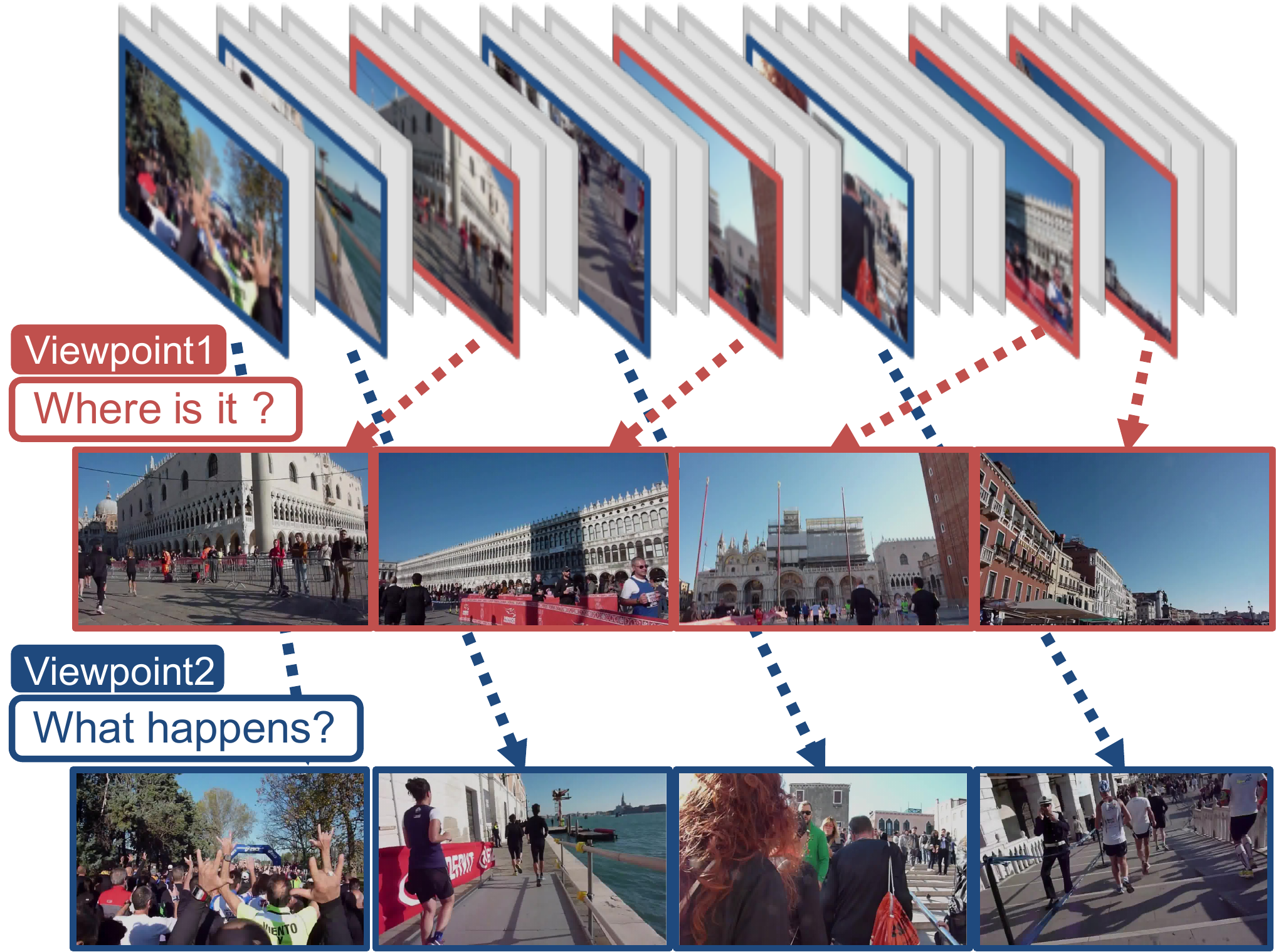}
\end{tabular}
\end{center}
\vspace{-0.6cm}
\caption{Many types of summaries can exist for one video based on the {\it viewpoint} toward it.}
\label{fig:motivation}
\vspace{-0cm}
\end{figure}

Owing to the recent spread of Internet services and inexpensive cameras, an enormous number of videos have become available, making it difficult to verify all content. 
Thus, video summarization, which compresses a video by extracting the {\it important} parts while avoiding redundancy, has attracted the attention of many researchers.

The information deemed {\it important} can be varied based on the particular aspect the viewer focuses on, which hereafter we will refer to as {\it viewpoint} in this paper\footnote[1]{Note it does not mean the physical position.}. For instance, given the video in which the running events take place in Venice, as shown in Fig.~\ref{fig:motivation}, if we watch it focusing on the ``kind of activity,'' the scene in which many runners come across in front of the camera is considered to be important. 
Alternatively, if the attention is focused on ``place,'' the scene that shows a beautiful building may be more important. Such {\it viewpoints} may not be limited to explicit ones stated in the above examples, and in this sense, the optimal summary is not necessarily determined in only one way. 

Most existing summarization methods, however, assume there is only one optimal for one video. Even though the variance between subjects are considered by comparing multiple human-created summaries during evaluation, it is difficult to determine how well the {\it viewpoint} is considered.

\begin{figure}[t]
\begin{center}
\begin{tabular}{c}
\includegraphics[clip, width=0.85\linewidth]{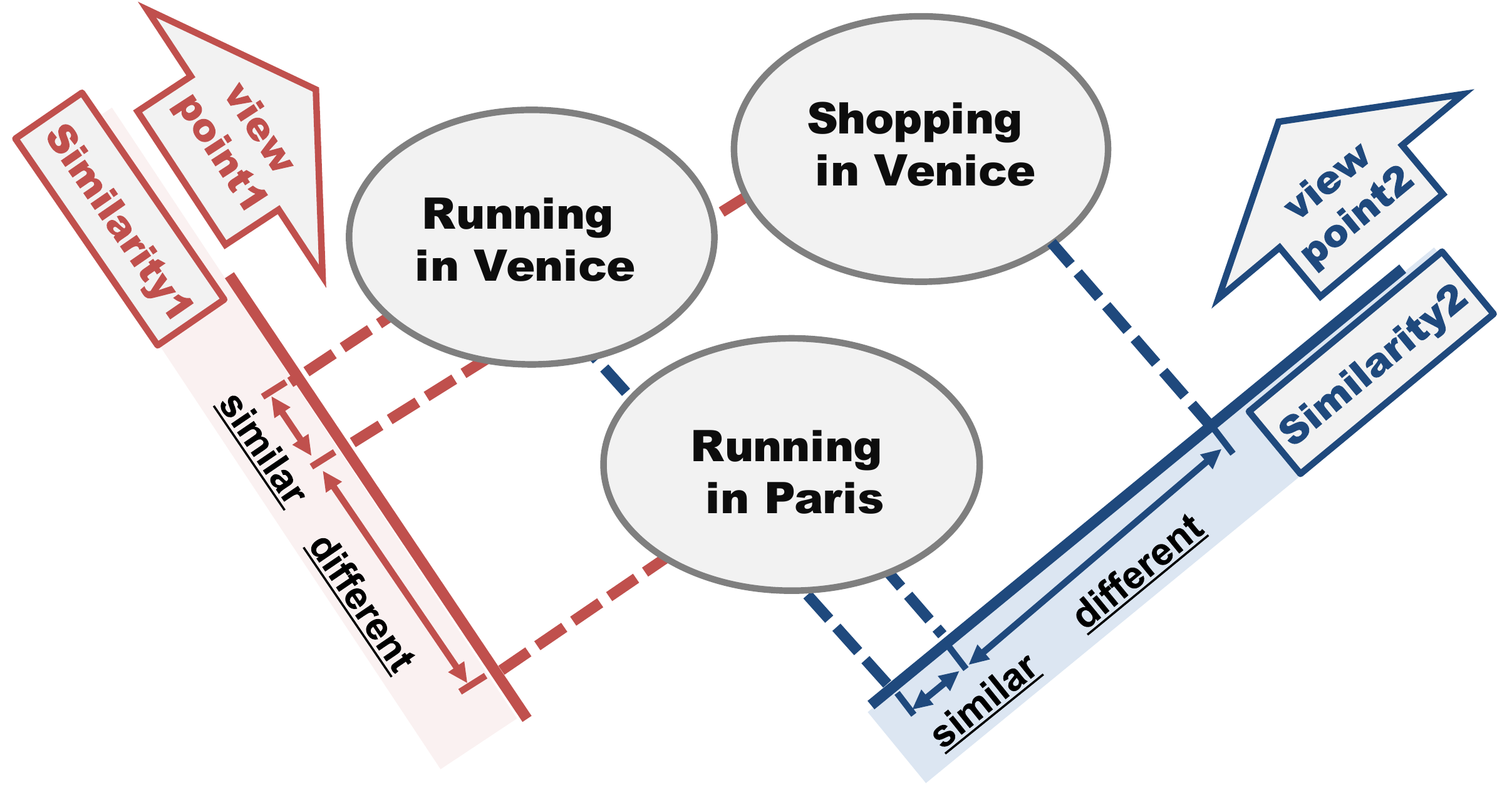}
\end{tabular}
\end{center}
\vspace{-0.7cm}
\caption{Conceptual relationship between a {\it viewpoint} and {\it similarity}. This paper assumes a {\it similarity} is derived from a corresponding {\it viewpoint}.}
\label{fig:concept}
\vspace{-0cm}
\end{figure}

Although several different ways may exist for interpreting a {\it viewpoint}, this paper takes the approach of dealing with it by considering the {\it similarity}, which represents what we feel is similar or dissimilar, and has a close relationship with the {\it viewpoint}.
For example, as shown in Fig.~\ref{fig:concept}, ``running in Paris'' is closer to ``running in Venice'' than ``shopping in Venice'' from the {\it viewpoint} of the ``kind of activity,'' but such a relationship will be reversed when the {\it viewpoint} changes to ``place.''
Here, we use the word {\it similarity} to indicate the one that captures semantic information rather than the appearance, and importantly, it is changeable depending on the {\it viewpoint}. We aim to generate a summary considering such {\it similarities}.
A natural question here is ``where does the {\it similarity} come from?''

We may be able to obtain it by asking someone whether two frames are similar or dissimilar for all pairs of frames (or short clips). Given that {\it similarity} changes depending on its {\it viewpoint}, it is unrealistic to obtain frame-level similarity for all {\it viewpoints} in this manner.

This paper particularly focuses on video-level {\it similarities}. More concretely, we utilize the information of how multiple videos are divided into groups as an indicator of {\it similarity} because of its accessibility. For example, we have multiple video folders on our PCs or smart-phones, or we sometimes categorize videos on an Internet service. They are divided according to a reason, but in most cases, why they are grouped the way they are is unknown, or irrelevant to criteria, such as preference (liked or not liked). 
Thus, a {\it viewpoint} is not evident, but such video-level {\it similarity} can be measured as a mapping of {\it one viewpoint}.

In this paper, we assume the situation that multiple groups of videos that are divided based on {\it one similarity} are given, and we investigate how to introduce unknown underlying {\it viewpoint} to the summary.
It is worth noting that, as we assume there are multiple possible ways to divide videos into groups depending on a {\it viewpoint} given the same set of videos, some overlap of content can exist between videos belonging to different groups, leading to technical difficulties, as we will state in Section~\ref{sec:relatedwork}.

For considering {\it similarity},
summaries extracted from similar videos should be similar, and ones extracted from different videos should be different from each other in addition to avoiding the redundancy derived from the original motivation of video summarization. 
In other words, given multiple groups of videos, the output summary of the video summarization algorithm should be: 
(A) diverse,
(B) representative of videos in the same group, and 
(C) discriminative against videos in the different groups.

To satisfy the requirements (A)-(C) simultaneously, we proposed a novel video summarization method from multiple groups of videos. Inspired by Fisher's discriminant criteria, it selects a summary by optimizing the combination of three terms the (a) inner-summary, (b) inner-group, and (c) between-group variance defined based on the feature representation of the summary, which can simply represent (A)-(C). 
In addition, we developed a novel optimization algorithm, which can be easily combined with feature learning, such as using convolutional neural networks (CNNs).

Moreover, we developed a novel dataset to investigate how well the generated summary reflects an underlying {\it viewpoint}. Because knowing individual {\it viewpoint} is generally impossible, we fixed it to two types of topics for each video. We also collected multiple videos that can be divided into groups based on these {\it viewpoints}.
Quantitative and qualitative experiments were conducted on the dataset to demonstrate the effectiveness of proposed method.

The contributions of this paper are as follows:
\begin{itemize}
\setlength{\parskip}{0.05cm}
\setlength{\itemsep}{0.05cm}
\item Propose a novel video summarization method from multiple groups of videos where their {\it similarity} are taken into consideration,
\item Develop a novel dataset for quantitative evaluation 
\item Demonstrate the effectiveness of proposed method by quantitative and qualitative experiments on the dataset.
\end{itemize}

The remainder of this paper is organized as follows. In Section~\ref{sec:relatedwork}, we discuss the related work of video summarization. Further, we explain the formulation and optimization of our video summarization method in Section~\ref{sec:proposed}. We state the detail of the dataset we created in Section~\ref{sec:dataset}, and describe and discuss the experiments that we performed on it in Section~\ref{sec:experiment}. 
Finally, we conclude our paper in Section~\ref{sec:conclusion}.

\section{Related work}\label{sec:relatedwork}
Many recent studies have tackled the video summarization problem, and most of them can be categorized into either unsupervised or supervised approach. Unsupervised summarization~\cite{ngo2003automatic, lu2013story, liu2002optimization, ma2002user, chen2011formulating, fleischman2007temporal, zhu2007trajectory, hong2009event,khosla2013large, kim2014joint, song2015tvsum, mahasseni2017unsupervised, elhamifar2017online} that creates a summary using specific selection criteria, has been conventionally studied. However, owing to the subjective property of this task, a supervised approach~\cite{lee2012discovering, sun2014ranking, potapov2014category, liu2010hierarchical, GygliECCV14,plummer2017enhancing, gygli2015video,kulesza2012determinantal,gong2014diverse,zhang2016summary}, that trains a summarization model which takes human-created summaries as the supervision, became standard because of its better performance.
Most of their methods aim to extract one optimal summary and do not consider the {\it viewpoint}, which we focus on in this study.

\begin{figure}[t]
\begin{center}
\begin{tabular}{c}
\hspace{-0.6cm}
\includegraphics[clip, width=\linewidth]{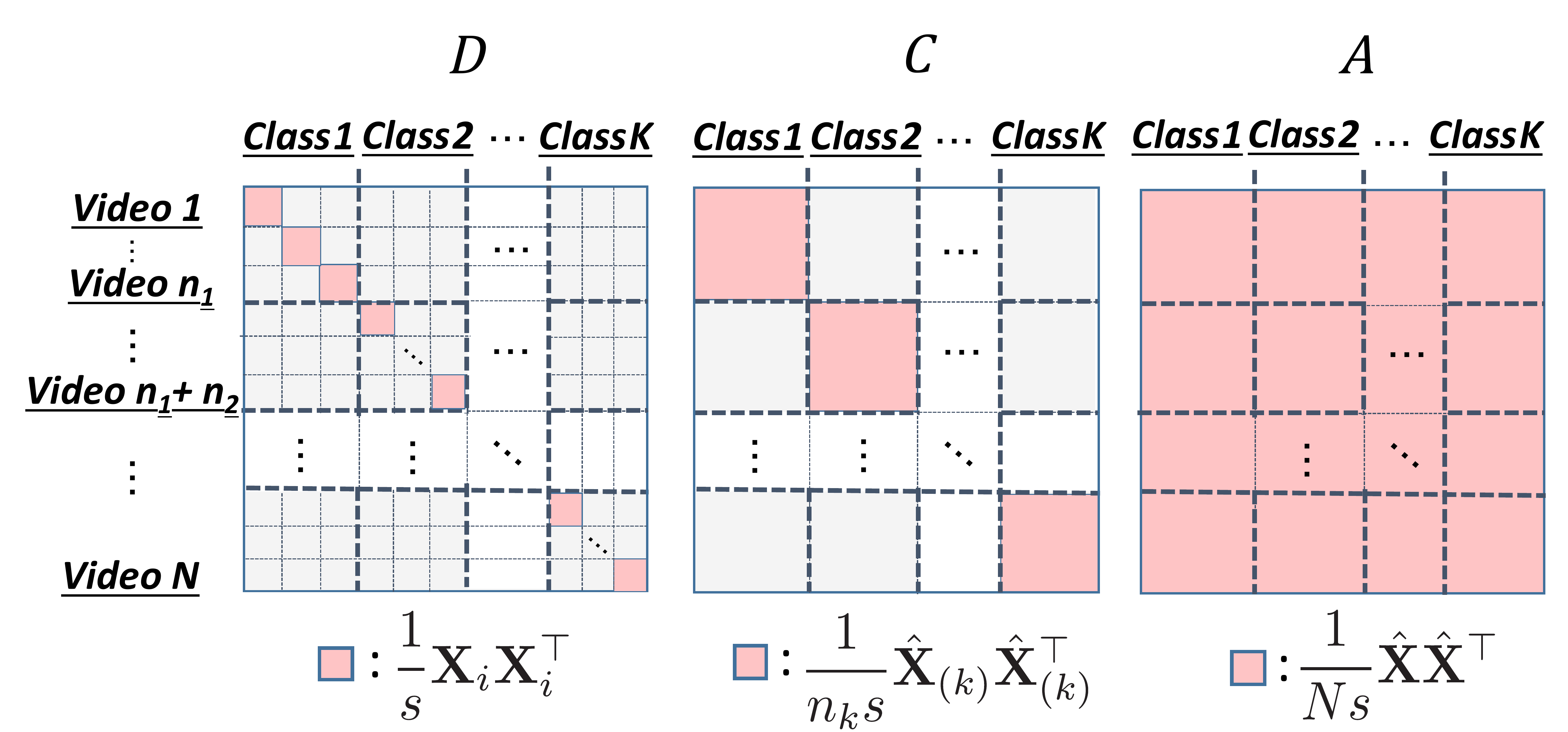}
\end{tabular}
\vspace{-0.6cm}
\end{center}
\caption{Overview of matrices D, C, and A, which are similarity matrices of inner-video, inner-group, and all videos. Non-zero elements of each matrix are colored pink and zero elements are colored gray. }
\label{fig:mat}
\vspace{-0cm}
\end{figure}

The exception is query extractive summarization~\cite{sharghi2016query, sharghi2017query} whose model takes a keyword as input and generates a summary based on it. It is similar to our work in that it assumes there can be multiple kinds of summaries for one video. However, our work is different in that we estimate what summary is created base on from the data instead of taking it as input. 
Besides, training model requires frame-level importance annotation for each keyword, which is unrealistic for real applications.

Some of the previous research worked on video summarization utilizing only other videos to alleviate the difficulty of building a dataset~\cite{chu2015video,Panda_2017_ICCV,Panda_2017_CVPR}.
\cite{chu2015video,Panda_2017_CVPR} utilized other similar videos and aims to generate a summary that is (A) diverse, and (B) representative of videos in a similar group, but it is not considered to be (C) discriminative against videos in different groups. Given that not only {\it what is similar} but also {\it what is dissimilar} is essential to consider {\it similarity}, we attempt to generate a summary that meets all of the conditions, (A)-(C).

The research most relevant to ours is~\cite{Panda_2017_ICCV}, which attempted to introduce discriminative information by utilizing a trained video classification model. It generates a summary with two steps. In the first step, it trains a spatio-temporal CNN that classifies the category of each video. In the second step, it calculates importance scores by spatially and temporally aggregating the gradients of the network's output with regard to the input over clips.

The success of this method has a strong dependence on the training in the first step. In this step, training is performed clip-by-clip by assigning the same label as that the video belongs to, to all clips of the video. Thus, it implicitly assumes all clips can be classified to the same group, and if there are some clips that are difficult to classify, it suffers from over-fitting caused by trying to classify it correctly. Such a strong assumption does not apply in general, because generic videos (such as ones on YouTube) include various types of content.
This assumption does not also apply in our case because we are interested in the situation where there are multiple possible ways to divide videos into groups given the same set of videos, as stated in Section~\ref{sec:introduction}, where some parts of videos can overlap with ones belonging to different groups for some {\it viewpoints}.

Unlike this, we do not assume all clips in the video can be classified correctly. Instead, our method considers the discrimination for only parts of videos. This makes it easy to find discriminative information even when there are visually similar clips across different groups.

We also acknowledge methods for discovering mid-level discriminative patches~\cite{singh2012unsupervised,li2015mid,jain2013representing,doersch2013mid,doersch2012makes,doersch2015makes} as related works because it attempts to find representative and discriminative elements from grouped data. Our work can be regarded as an extension of them to general videos.

\section{Method}\label{sec:proposed}
First, we introduce three quantities, that is, the (a) inner-summary, (b) within-group, and (c) between-group variances in subsection~\ref{subsec:formulation}. Subsequently, we formulate our method by defining a loss function to meet the requirements discussed in Section~\ref{sec:introduction}. The optimization algorithm is described in subsection~\ref{subsec:optimization}, and how to combine it with CNN feature learning is mentioned in subsection~\ref{subsec:fl}.
The detailed derivation can be found in the supplemental material.

\subsection{Formulation}\label{subsec:formulation}
Let ${\mathbf X}_{i} = [{\mathbf x}_1, {\mathbf x}_2, ..., {\mathbf x}_{T_{i}}]^{\top} \in {\mathbb R}^{T_{i}  \times d}$ be a feature matrix for a video $i$ with $T_{i}$ segment (or frame) features ${\mathbf x}$. Our goal is to select $s$ segments from the video. 
We start by defining the feature representation of the summary for video $i$ as ${\mathbf v}_i=\frac{1}{s}{\mathbf X}_{i}^{\top}{\mathbf z}_i$, where ${\mathbf z}_i \in \{0, 1\}^{T_{i}}$ is the indicator variable and $z_{it}=1$ if the $t$-th segment is selected, and otherwise 0. It also has a constraint $||{\mathbf z}_i||_{0}=s$ indicating that just $s$ segments are selected as a summary. 
We can define a variance $S_i^V$ for the summary of a video $i$ as
\begin{eqnarray}
S_{i}^V &=& \sum_{t=1}^{T_{i}} z_t({\mathbf x}_t - {\mathbf v}_{i})({\mathbf x}_t - {\mathbf v}_{i})^{\top}.
\end{eqnarray} 
Thus, its trace can be written as:
\begin{eqnarray}
Tr(S_{i}^V) 
&=&  \sum_{t=1}^{T_{i}} z_t{\mathbf x}_t^{\top}{\mathbf x}_t - \frac{1}{s}{\mathbf z}_{i}^{\top} {\mathbf X}_{i}{\mathbf X}_{i}^{\top}{\mathbf z}_{i}.
\end{eqnarray}
Placing all $N$ videos together by using a stacked variable ${\hat {\mathbf z}}= [ {\mathbf z}^{\top}_1, {\mathbf z}^{\top}_2, ..., {\mathbf z}^{\top}_N  ]^{\top} \in \{0, 1\}^{\sum_{i=1}^{N} T_i}$, we can rewrite
\begin{eqnarray}\label{eq:video_var}
Tr(S^V)&=&\sum_{i=1}^{N} Tr( S_{i}^V) = {\hat {\mathbf z}}^{\top} (F - D){\hat {\mathbf z}}.
\end{eqnarray}
where $F$ is a diagonal matrix whose element corresponds to ${\mathbf x}_{t}^{\top}{\mathbf x}_{t}$, and $D =  \frac{1}{s} \oplus \sum_{i=1}^{N} {\mathbf X}_{i}{\mathbf X}_{i}^{\top}$ is a block diagonal matrix containing a similarity matrix of segments in the video $i$ as $i$-th block elements.

By exploiting categorical information, we can also compute within-group variance $S^W$ and between-group variance $S^B$. To compute them, we define the mean vector ${\bm \mu}_{k}$ for group $k \in \{1:K\}$ and global mean vector ${\bar {\bm \mu}}$ as: 
\begin{eqnarray}
{\bm \mu}_{k} &=& \frac{1}{n_k}\sum_{i \in L_{(k)}}{\mathbf v}_i = \frac{1}{n_{k}s}{\hat {\mathbf X}}_{(k)}^{\top} {\hat {\mathbf z}}_{(k)},\\
{\bm {\bar \mu}}&=& \frac{1}{N}\sum^{N}_{i=1}{\mathbf v}_i = \frac{1}{Ns} {\hat {\mathbf X}}^{\top} {\hat {\mathbf z}},
\end{eqnarray}

\begin{algorithm}[t]
\caption{Optimization algorithm of (\ref{eq:opt_relax})}\label{alg:opt}
\begin{algorithmic}[1]
\State{\bf INPUT}: data matrix $Q=Q_1-Q_2$, the number of selected clips $s$.
\State {\bf INITIALIZE}: ${\mathbf z}_{i}=(1/s)\ \mathbf{1}_{T_i}$ for all video index $i$. 
\Repeat 
\State Calculate upper bound ${\hat L}_{(t)} = {\hat {\mathbf z}}^{\top}Q_{1} {\hat {\mathbf z}} - 2\  {\hat {\mathbf z}}_{(t)}^{\top}Q_{2} {\hat {\mathbf z}}$
\State Replace loss with ${\hat L}_{(t)}$ and solve QP problem.
\Until{convergence}
\State {\bf RETURN $ {\hat {\mathbf z}}$}
\end{algorithmic}
\end{algorithm}
respectively. In these equations, $L_{(k)}$ is the set of indices of videos belonging to group $k$ and $n_k=|L_{(k)}|$ (i.e., $N=\sum_k n_k$).
In addition, ${\hat {\mathbf X}}=[{\mathbf X}_{1}^{\top}|{\mathbf X}_{2}^{\top}|...|{\mathbf X}_{N}^{\top}]^{\top} \in {\mathbb R}^{(\sum_{i=1}^{N} T_i)\times d}$ is the matrix stacking all segment features of all videos. ${\hat {\mathbf X}_{(k)}}$ and ${\hat {\mathbf z}_{(k)}}$ are parts of ${\hat {\mathbf X}}$ and ${\hat {\mathbf z}}$, respectively, corresponding to videos contained by group $k$. We assume that a video index is ordered to satisfy ${\hat {\mathbf X}} = [{\hat {\mathbf X}_{(1)}}^{\top}|{\hat {\mathbf X}_{(2)}}^{\top}|...|{\hat {\mathbf X}_{(K)}}^{\top}]^{\top}$ .
Here, the trace of within-group variance for group $k$ can be written as:
\begin{eqnarray}
Tr(S^{W}_{(k)})=Tr(\sum_{i \in L_{(k)}} s ({\mathbf v}_i-{\bm \mu}_k)({\mathbf v}_i-{\bm \mu}_k)^{\top}) \nonumber \\
= \frac{1}{s}  \sum_{i \in L_{(k)}} {\mathbf z}_{i}^{\top} {\mathbf X}_{i}{\mathbf X}_{i}^{\top}{\mathbf z}_{i} - \frac{1}{n_{k}s} {\hat {\mathbf z}}_{(k)}^{\top} {\hat {\mathbf X}}_{(k)}{\hat {\mathbf X}}_{(k)}^{\top}{\hat {\mathbf z}}_{(k)}.
\end{eqnarray}
Aggregating them over all groups, the trace of within-group variance takes the following form:
\begin{eqnarray}\label{eq:wicls_var}
Tr(S^{W})=\sum_{k=1}^{K} Tr(S^{W}_{(k)}) = {\hat {\mathbf z}}^{\top} (D - C) {\hat {\mathbf z}}.
\end{eqnarray}
$C = \frac{1}{s} \oplus \sum_{k=1}^{K} \frac{1}{n_k}{\hat {\mathbf X}_{(k)}}{\hat {\mathbf X}_{(k)}}^{\top}$ is a block diagonal matrix containing a similarity matrix of segments in the video belonging to group $k$ as a $k$-th block element.
Similarly, the trace of between-group variance is: 
\begin{eqnarray} \label{eq:btcls_var}
Tr(S^B) &=& Tr(\sum_{k=1}^{K} n_k s ({\bm \mu}_k-{\bar {\bm \mu}}) ({\bm \mu}_k-{\bar {\bm \mu}})^{\top})\nonumber \\
&=& {\hat {\mathbf z}}^{\top} (C - A) {\hat {\mathbf z}}.
\end{eqnarray}
In addition, matrix $A$ is defined by $A  = \frac{1}{Ns} {\hat {\mathbf X}} {\hat {\mathbf X}}^{\top}$.
We show the overview of matrices $D$, $C$, and $A$ in Fig.~\ref{fig:mat}. 
\\

{\bf Loss function}: We designed an optimization problem to meet the requirements discussed in Section~\ref{sec:introduction}:
(A) diverse,
(B) representative of videos in the same group, and 
(C) discriminative against videos in different groups.
To simultaneously satisfy them, we minimized the within-group variance while maximizing the between-group and inner-video variances inspired by the concept of linear discriminant analysis. Thus, we maximized the following function, which is the weighted sum of the aforementioned three terms:
\begin{eqnarray}\label{eq:loss}
&\lambda_1 Tr(S^V) - \lambda_2 Tr(S^W) + \lambda_3Tr(S^B) &\nonumber \\
&{\rm s.t.}\ \ \lambda_1 \geq 0,\ \ \lambda_2 \geq 0, \ \  \lambda_3 \geq 0,&
\end{eqnarray}
where $\lambda_1$, $\lambda_2$, $\lambda_3$ are hyper-parameters that control the importance of each term. We empirically fixed $\lambda_1=0.05$ in our experiments.

By substituting (\ref{eq:video_var}), (\ref{eq:wicls_var}), and (\ref{eq:btcls_var}) into (\ref{eq:loss}), the optimization problem can be solved as: 
\begin{eqnarray} \label{eq:opt}
&{\min}& {\hat {\mathbf z}}^{\top}Q{\hat {\mathbf z}}\nonumber \\
&Q \triangleq& - \lambda_{1} F + (\lambda_1 + \lambda_2)D - (\lambda_2 + \lambda_3)C + \lambda_3 A\nonumber \\
 &{\rm s.t.}& ||{\mathbf z}_i||_{0}=s, \ {\mathbf z}_i \in \{0, 1\}^{T_{i}}, \ \forall i \in \{1:N\}
\end{eqnarray}

\begin{table*}[t!]
\small 
  \centering
  \caption{The list of names for video groups (\textbf{target group}, \textbf{related group1}, \textbf{related group2}), and individual concepts of \textbf{target group} (\textbf{concept1}, \textbf{concept2}). We omit the article (e.g., the) before nouns due to the lack of space. We use the abbreviation of target group as [RV, RB, BS, DS, RD, SR, CC, RN, SC, RS] from top to bottom.}
  \vspace{-0.35cm}
  \label{tab:table1}
  \bgroup
  \def\arraystretch{1.25}
   \begin{tabular}{|c||c|c|c|c|} \hline 
    \textbf{target group (TG)}&\textbf{concept1}&\textbf{concept2}&\textbf{related group1 (RG1)}&\textbf{related group2 (RG2)}\\ \hline \hline
    running in Venice&Venice&running&running in Paris&shopping in Venice\\ \hline
    riding bike on beach&beach&riding bike&riding bike in city&surfing on beach\\ \hline
   boarding on snow mountain&snow mountain&boarding&boarding on dry sloop&hike in snow mountain\\ \hline 
    dog chasing sheep&sheep&dog&dog playing with kids&sheep grazing grass\\ \hline
    racing in desert&desert&racing&racing in circuit&riding camel in desert\\ \hline
    swimming and riding bike&swimming&riding bike&riding bike and tricking&diving and swimming\\ \hline
   catching and cooking fish&catching fish&cooking fish&cooking fish in village&catching fish at river\\ \hline
   riding helicopter in NewYork&NewYork&helicopter&riding helicopter in Hawaii&riding ship in NewYork\\ \hline
   slackline and rock climbing&slackline&rock climbing&rock climbing and camping&slcakline and jaggling\\ \hline
   riding horse in safari&safari&riding horse&riding horse in mountain&riding vehicle in safari\\ \hline
  \end{tabular}
  \egroup
  \vspace{-0.3cm}
\end{table*}

\subsection{Optimization} \label{subsec:optimization}
Given that minimizing (\ref{eq:opt}) directly is infeasible, we relaxed it to a continuous problem as follows:
\begin{eqnarray} \label{eq:opt_relax}
&{\min}& {\hat {\mathbf z}}^{\top}Q{\hat {\mathbf z}}\nonumber \\
 &{\rm s.t.}& P {\hat {\mathbf z}}=s{\mathbf 1}_{N}, \  {\hat {\mathbf z}}\in [0, 1]^{\sum_{i=1}^{N} T_i}\nonumber \\ 
  &{\rm where}&P^{\top}=\left[
\begin{array}{cccc}
  {\mathbf 1}_{T_1} &{\mathbf 0}&\cdots&{\mathbf 0}\\
{\mathbf 0} &{\mathbf 1}_{T_2}&\cdots&\vdots \\
\vdots &\vdots &\ddots&\vdots \\
 {\mathbf 0} &{\mathbf 0}&\cdots&{\mathbf 1}_{T_N}\\
\end{array}
\right].
\end{eqnarray}
${\mathbf 1}_a$ indicates a vector whose elements are all ones and whose size is $a$, and the size of matrix $P$ is $N\times \sum_{i=1}^{N} T_i$. The designed optimization problem is the difference of convex (DC) programming problem because all matrices that compose $Q$ in (\ref{eq:opt_relax}) are positive semi-definite. We utilized a well-known CCCP (concave convex procedure) algorithm~\cite{yuille2002concave, yuille2003concave} to solve it. 
Given the loss function represented by $L({\mathbf x}) = f({\mathbf x})-g({\mathbf x})$ where $f(\cdot)$ and $g(\cdot)$ are convex functions, the algorithm iteratively minimizes the upper bound of loss calculated by the linear approximation of $g({\mathbf x})$. Formally, in the iteration $t$, it minimizes: ${\hat L({\mathbf x})} = f({\mathbf x}) - {\partial}_{\mathbf x} g({\mathbf x}_{(t)})^{\top} {\mathbf x} \ge L({\mathbf x})$. In our problem, the loss function can be decomposed into the difference of two convex functions:
${\hat {\mathbf z}}^{\top}Q {\hat {\mathbf z}} =  {\hat {\mathbf z}}^{\top}Q_{1} {\hat {\mathbf z}} -  {\hat {\mathbf z}}^{\top}Q_{2} {\hat {\mathbf z}}$, where $Q_{1} \triangleq  (\lambda_1 + \lambda_2)D + \lambda_3 A$ and $Q_{2} \triangleq \lambda_{1} F  + (\lambda_2 + \lambda_3)C$. We optimized the following quadratic programming (QP) problem in $t$-th iteration,
\begin{eqnarray}\label{eq:qp}
&{\rm min}&{\hat {\mathbf z}}^{\top}Q_{1} {\hat {\mathbf z}} - 2\  {\hat {\mathbf z}}_{(t)}^{\top}Q_{2} {\hat {\mathbf z}}\nonumber \\
 &{\rm s.t.}& P {\hat {\mathbf z}}=s{\mathbf 1}_{N}, \  {\hat {\mathbf z}}\in [0, 1]^{\sum_{i=1}^{N} T_i},
\end{eqnarray}
where ${\hat {\mathbf z}}_{(t)}$ is the estimation of ${\hat {\mathbf z}}$ in the $t$-th iteration.
In our implementation, we used a CVX package~\cite{cvx, gb08} to solve the QP problem (\ref{eq:qp}).
An overview of our algorithm is shown in Algorithm~\ref{alg:opt}.
Please refer \cite{lanckriet2009convergence} for the convergence property of CCCP.

\subsection{Feature learning}\label{subsec:fl}
To obtain the feature representation that is more suitable for video summarization, feature learning is applied. 
Firstly, we replace the visual feature $\mathbf x$ in subsection~\ref{subsec:formulation} to $f({\mathbf x}; {\mathbf w})$ where $f(\cdot)$ is a feature extractor function that is differentiable with regard to the parameter ${\mathbf w}$ and the input ${\mathbf x}$ is a sequence of raw frames in the RGB space. Specifically, we exploited the C3D network~\cite{tran2015learning} as a feature extractor.
Fixing ${\hat {\mathbf z}}$, the loss function  (\ref{eq:opt_relax}) can be written as:
\begin{eqnarray} \label{eq:loss_fl}
L = \sum_{i, j} {\hat z}_i {\hat z}_j m_{ij} f({\mathbf x}_i)^{\top} f({\mathbf x}_j),
\end{eqnarray}
where ${\hat z}_i$ is $i$-th element of ${\hat {\mathbf z}}$. Also, $m_{ij}$ is the $ij$-th element of matrix $M$ written as follows:
\begin{eqnarray}
M = - \lambda_{1}{\mathbf 1}_{F} + (\lambda_1 + \lambda_2){\mathbf 1}_{D} - (\lambda_2 + \lambda_3){\mathbf 1}_{C} + \lambda_3 {\mathbf 1}_{A}. \nonumber
\end{eqnarray}
Here, ${\mathbf 1}_{X}$ represents an indicator matrix whose element takes 1 where the corresponding element of $X$ is not 0, and takes 0 otherwise.
We optimize the loss function with regard to the parameter by stochastic gradient decent (SGD). 
Because many of ${\hat z}_i$ are small values or zeros, minimizing (\ref{eq:loss_fl}) directly is not efficient. We avoid the inefficiency by sampling samples ${\mathbf x}_i$ based on their weight ${\hat z}_i$. Given $\sum {\hat z}_i=Ns$, we sample ${\mathbf x}_i$ from the distribution $p({\mathbf x}_i) =  {\hat z_i} / Ns \ (\ge 0)$ and stochastically minimize the expectation:
\begin{eqnarray} \label{eq:loss_fl_exp}
{\mathbb E}_{{\mathbf x}_{i}, {\mathbf x}_{j} \sim p({\mathbf x})}[m_{ij} f({\mathbf x}_i)^{\top} f({\mathbf x}_j)].
\end{eqnarray}
In an iteration when updating parameters, the model fetches pairs $({\mathbf x}_{i}, {\mathbf x}_{j})$ and computes the dot product of the feature representations. The loss for this batch is calculated by summing up the dot product weighted by $m_{ij}$. 
We repeatedly and alternately compute the summary via the Algorithm~\ref{alg:opt} and optimize the parameter of the feature extractor.

\begin{table}[t!]
\vspace{-0.3cm}
\small 
  \centering
  \caption{statistics of dataset}
  \vspace{-0.25cm}
  \label{tab:table-statistics}
   \begin{tabular}{|c||c|c|c|} \hline 
    group & \# of videos &\# of frames  & duration \\ \hline\hline 
    TG & 50 & 243,873 & 8,832(s) \\ \hline 
    RG1 + RG2 & 100 & 440,330 & 15,683(s)  \\ \hline 
   \end{tabular}
   \vspace{0.1cm}
   \label{tb:statistics}
\end{table}

\begin{figure*}[t!]
    \centering 
\begin{subfigure}{0.49\textwidth}
  \includegraphics[width=\linewidth]{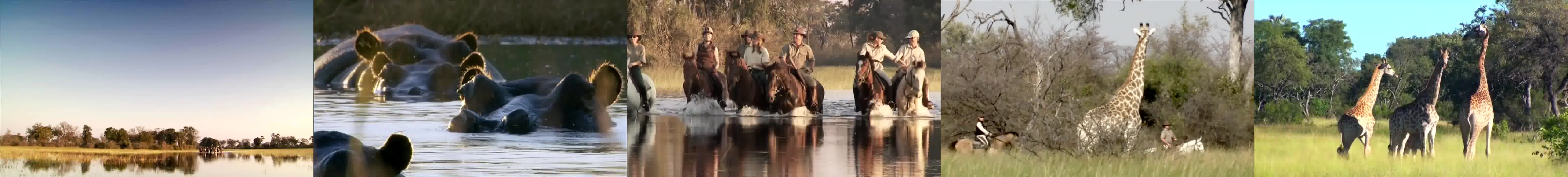}\vspace{-0cm}\\
\includegraphics[width=\linewidth]{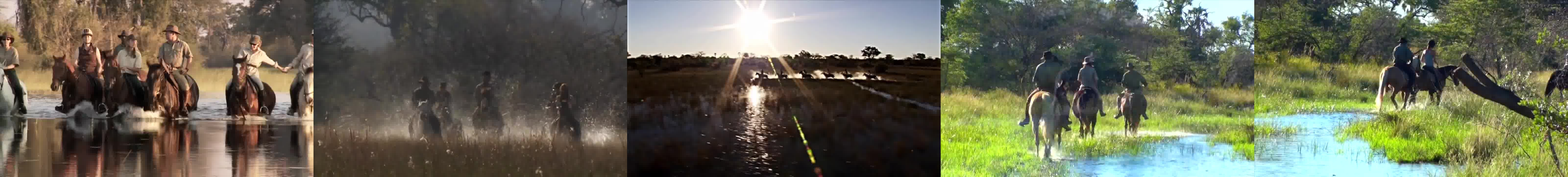}\subcaption{safari (above) and riding horse (below)}
  \label{fig:1}
\end{subfigure}\hfil
\begin{subfigure}{0.49\textwidth}
  \includegraphics[width=\linewidth]{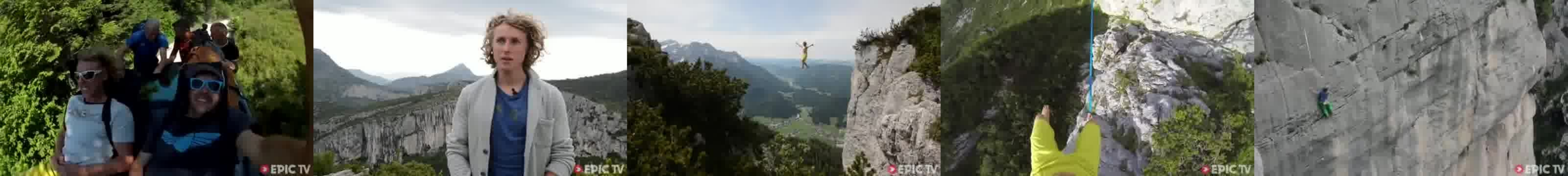}\vspace{0.0cm}\\
\includegraphics[width=\linewidth]{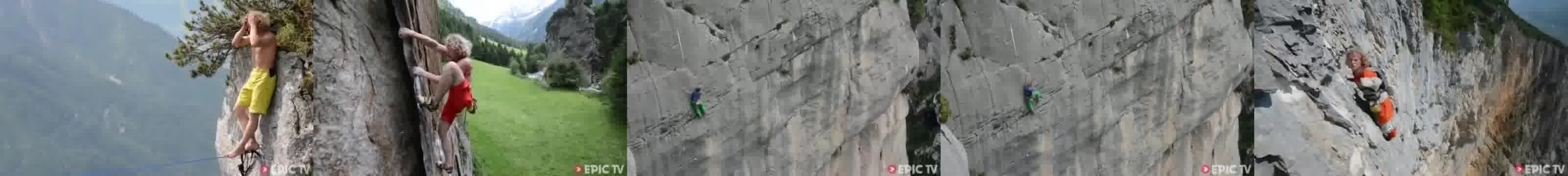}\subcaption{slackline (above) and rock climbing (below)}
\end{subfigure}\\
 \centering 
\begin{subfigure}{0.49\textwidth}
  \includegraphics[width=\linewidth]{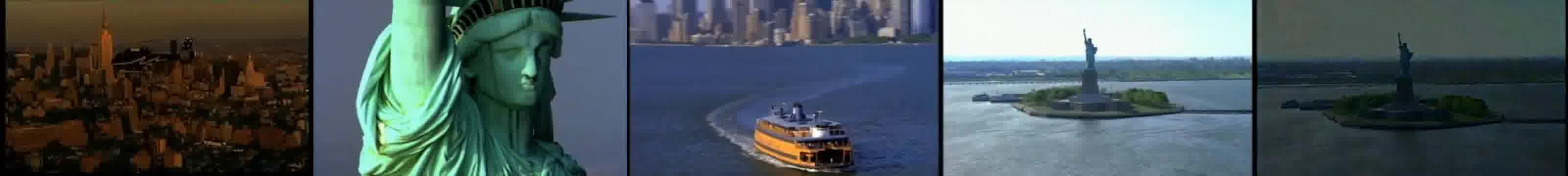}\vspace{0.02cm}\\
\includegraphics[width=\linewidth]{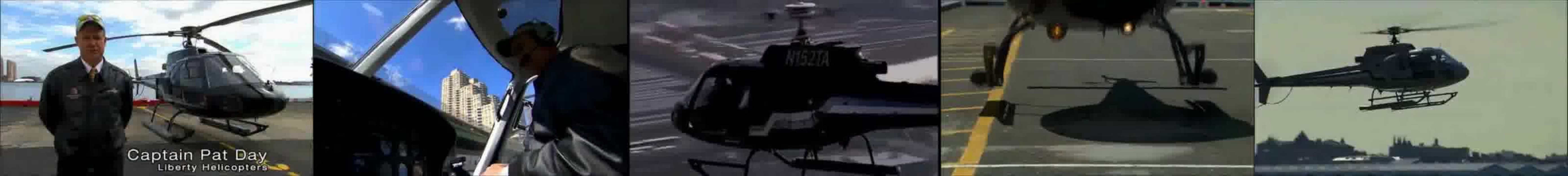}\subcaption{NewYork (above) and riding helicopter (below)}
\end{subfigure}
\begin{subfigure}{0.49\textwidth}
  \includegraphics[width=\linewidth]{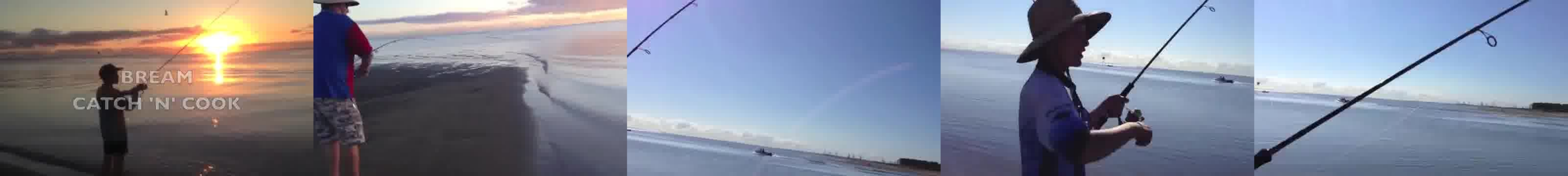}\vspace{0.02cm}\\
\includegraphics[width=\linewidth]{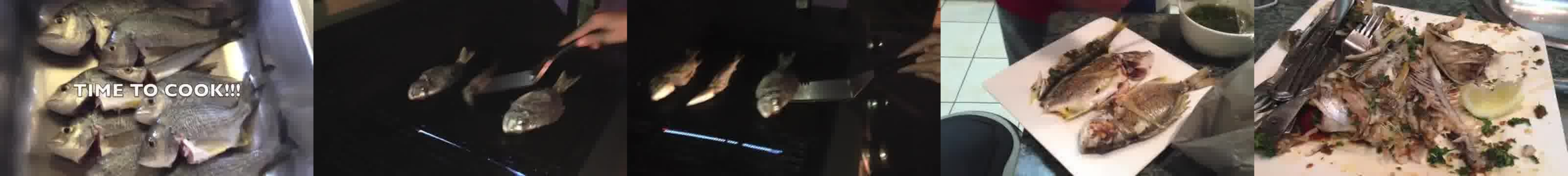}
\subcaption{catching fish (above) and cooking fish (below)}
\end{subfigure}
 \vspace{-0.3cm}
   \caption{Example human-created summary of video whose \textbf{target group} are ``riding horse in safari'' (upper left), ``slackline and rock climbing'' (upper right), ``riding helicopter in NewYork'' (lower left), and ``catching and cooking fish'' (lower right) based on the concept written in each figure.}
      \vspace{-0.6cm}
  \label{fig:example images}
\end{figure*}

\begin{figure}[t!]
\vspace{0.2cm}
\begin{center}
\begin{tabular}{c}
\includegraphics[clip, width=0.98\linewidth]{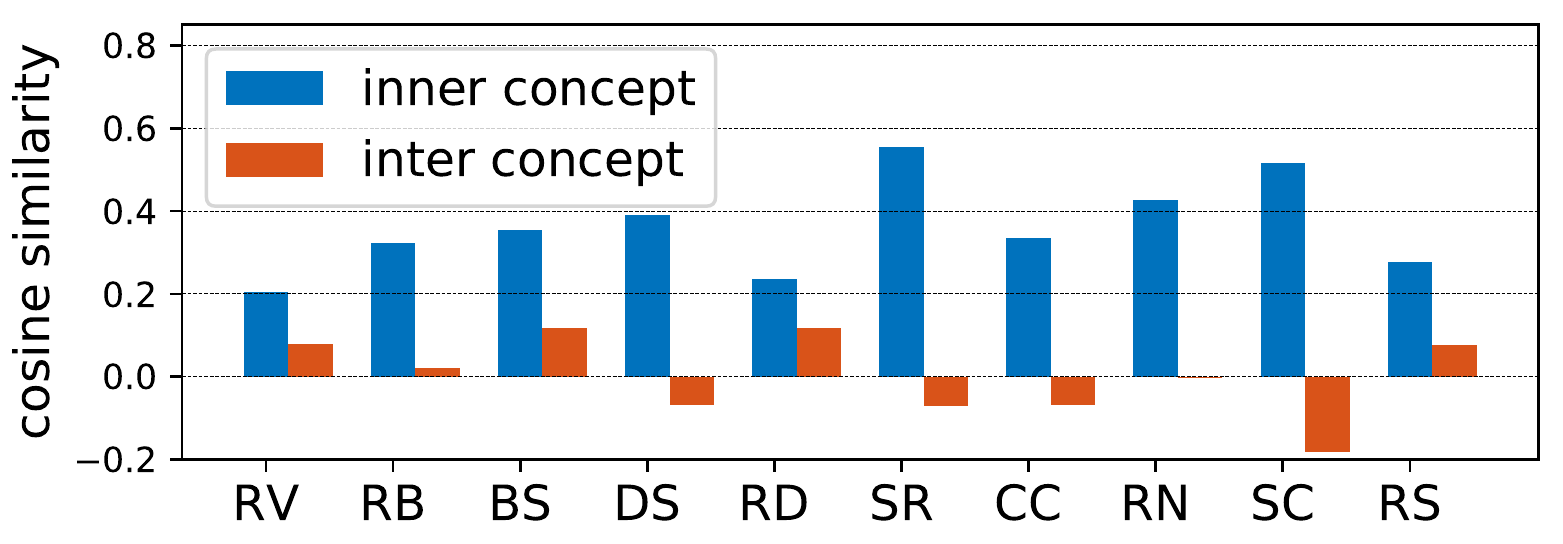}
\end{tabular}
\vspace{-0.4cm}
\caption{Mean cosine similarity of human-assigned scores for each target group. We denote the value computed from the score pairs that are assigned to the same concept and different concepts as inner concepts (blue) and inter concepts (orange), respectively. When referring to the abbreviated names of groups, please refer to the Table~\ref{tab:table1}.}
\vspace{-0cm}
\label{fig:label-similarity}
\end{center}
\end{figure}

\section{Dataset}\label{sec:dataset}
The motivation of this study is the claim that an optimal summary should be varied depending on a {\it viewpoint}, and this paper deals with this by considering the {\it similarities}.
To investigate how well the underlying {\it viewpoint} are taken into consideration, given multiple groups of videos that are divided based on the {\it similarity}, we compiled a novel video summarization dataset\footnote[2]{Dataset is available at https://akanehira.github.io/viewpoint/.}.
Quantitative evaluation is challenging because the {\it viewpoint} is generally unknown. Thus, for the purpose of quantitative evaluation, we collected a set of videos that can have two interpretable ways of separation assuming they have corresponding {\it viewpoint}. In addition, we collected human-created summaries fixing the importance criteria to two concepts based on each {\it viewpoint}. 
The procedure of building the dataset is as follows:

First, we collected five videos that match the topics written in \textbf{target group (TG)}, \textbf{related group1 (RG1)}, \textbf{related group2 (RG2)} of  Table~\ref{tab:table1} by retrieving them in YouTube\footnote[3]{https://www.youtube.com/} using a keyword. 
Each of \textbf{TG}, \textbf{RG1}, \textbf{RG2} has two explicit concepts such that they can be visually confirmed; e.g., location, activity, object, and scene. 
The concepts of \textbf{TG} are written in \textbf{concept1} and \textbf{concept2} columns in the table, and both \textbf{RG1} and \textbf{RG2} were chosen to share either one of them. 
There are two interpretable ways to divide these sets of videos, i.e., (\textbf{TG} + \textbf{RG1}) vs. (\textbf{RG2}) and (\textbf{TG} + \textbf{RG2}) vs. (\textbf{RG1}) because \textbf{RG1} and \textbf{RG2} share one topic with \textbf{TG}. Assuming these divisions are based on one {\it viewpoint}, we collected the summary based on it using two concepts for videos belonging to \textbf{TG}.  
For example, if we are given two groups, one of which contains ``running in Venice'' and ``running in Paris'' videos, and the other group includes ``shopping in Venice'' videos, the underlying {\it viewpoint} is expected to be ``kind of activity.'' For such a scenario, we collected summaries based on ``running'' for the videos of ``running in Venice.''  

For annotating the importance of each frame of the video belonging to \textbf{TG}, we used Amazon Mechanical Turk (AMT). Firstly, videos were evenly divided into clips beforehand so that the length of each clip was two seconds long following the setting of \cite{song2015tvsum}. 
Subsequently, after workers watched a whole video, they were asked to assign a importance score to each clip of the video, assuming that they created a summary based on a pre-determined topic, which corresponds to the concept written in \textbf{concept1} or \textbf{concept2} columns in the Table~\ref{tab:table1}. 
Importance scores are chosen from 1 (not important) to 3 (very important), and workers were asked to guarantee the number of clips having a score of 3 falls in the range between 10\% and 20\% of the total number of clips in the video. For each video and each \textbf{concept}, five workers were assigned. 

We display the statistics of the dataset and some example of the human-created summary in Table~\ref{tb:statistics} and Fig.~\ref{fig:example images}, respectively. 
Also, in order to investigate how similar the assigned score between subjects is, we calculated the similarity of the score vector. After subtracting the mean value from each score, the mean cosine similarity for the pair of scores that are assigned for the same concepts (e.g., \textbf{concept1} and \textbf{concept1}) and different concepts (\textbf{concept1} and \textbf{concept2}) were separately computed, and the result is shown in Fig.~\ref{fig:label-similarity}. 
As we can see in the table, the similarity of scores that comes from the inner-concept is higher than that of inter-concept, which indicates that the importance depends on the {\it viewpoint} of the videos.

\section{Experiment}\label{sec:experiment}
\subsection{Preprocessing}
To compute the segment used as the smallest element for video summarization, we followed a simple method proposed in \cite{chu2015video}. After counting the difference of two consecutive frames in the RGB and HSV space, the points on which the total amount of change exceeds 75\% of all pixels were regarded as change points. Subsequently, we combined short clips into the following clip and evenly divided the long clips in order such that the number of frames in each clip was more than 32 and less than 112.

\subsection{Visual features}
For obtaining frame-level visual features, we exploited the intermediate state of the C3D~\cite{tran2015learning} network, which is known to be so generic that it can be used for other tasks, including video summarization~\cite{Panda_2017_CVPR}. 
We extracted the features from an fc6 layer of a network pre-trained on a Sports1M~\cite{karpathy2014large} dataset.
The length of the input was 16 frames, and features were extracted every 16 frames. The dimension of the output feature vector was 4,096. Clip-level representations were calculated by performing an average pooling over all frame-level features in each clip followed by a $l_2$ normalization. 

\subsection{Evaluation}
For a quantitative evaluation, we compared automatically generated summaries with human made ones.
First, we explain the grouping setting of videos.
There are two interpretable ways of grouping that include each target group as stated in Section~\ref{sec:dataset}:
\begin{itemize}[noitemsep,topsep=2pt]
\item regarding \textbf{related group2 (RG2)} as the same group as \textbf{target group (TG)} and \textbf{related group1 (RG1)} as the different group (setting1).
\item regarding \textbf{related group1 (RG1)} as the same group as \textbf{target group (TG)} and \textbf{related group2 (RG2)} as the different group (setting2).
\end{itemize}
In the case that the grouping setting1 was used, we evaluated it with the summary annotated for \textbf{concept1}. Alternatively, when videos are divided like setting2, the summary for \textbf{concept2} was used for the evaluation. Note we treated each \textbf{TG} independently in throughout this experiment. 

We set the ground-truth summary in the following procedure.
The mean of the importance scores were calculated over all frames in each clip, which was determined by the method described in the previous subsection. The top-$30\%$ of the number of all clips whose importance scores are highest were extracted from each video and regarded as ground-truth. 
As an evaluation metric, we computed the mean Average Precision (MAP) from a pair of summaries, and reported the mean value. Formally, for each \textbf{TG}, 
$1 / (CIJ) \sum^{C}_{c=1}\sum^{J}_{j=1}\sum^{I}_{i=1} AP(l^{ij}_{(c)}, {\hat l}^{i}_{(c)})$
was calculated where $l$ and ${\hat l}$ are ground-truth summaries and the predicted summary, respectively. $C$ indicates the number of concepts on which the summary created by the annotators is based on.
$I, J$ are the number of subjects and the number of videos in the group respectively. In particular, $(C, I, J)$ were (2, 5, 5) as written in Section~\ref{sec:dataset} in this study.

\subsection{Implementation detail}
As stated in Section~\ref{sec:proposed}, we used a C3D network~\cite{tran2015learning} pre-trained on a Sports1M dataset~\cite{karpathy2014large}, which has eight convolution layers followed by three fully connected layers. During fine-tuning, the initial learning rate was $10^{-5}$. Weight decay and momentum were set to $10^{-4}$ and $0.9$ respectively. The number of repetitions of the feature learning and summary estimation was set to 5. The number of epochs for each repetition was 10, and the learning rate was multiplied by 0.9 for every epoch. Here, epoch indicates \{\# of all clips\}/\{batch size\} iteration even though clips were not uniformly sampled.

\subsection{Comparison with other methods}
To investigate the effectiveness of the proposed method, we compared it with other baseline methods as follows:

{\bf Sparse Modeling Representative Selection (SMRS) ~\cite{elhamifar2012see}}: SMRS computes a representation of video clips such that a small number of clips can represent an entire video by group sparse regularization.
We selected clips whose $l_2$ norm of representation was the largest.

{\bf kmeans (CK) and spectral clustering (CS)}: 
One simple solution to extract representative information between multiple videos is applying clustering algorithm. We applied two clustering algorithms, namely kmeans (CK) and spectral clustering (CS), for all clips of video which was regarded as the same groups. RBF kernel was used to build an affinity matrix necessary for computation of spectral clustering.
The number of clusters was set to 20 as in \cite{Panda_2017_ICCV}. Summaries were generated by selecting clips that are the closest to the cluster center of the largest clusters. 

{\bf Maximum Bi-Clique Finding (MBF)} \cite{chu2015video}: The MBF is a video co-summarization algorithm that extracts a bi-clique from a bi-partite graph with a maximum inner weight. MBF algorithms were applied to each pair of videos within a video group, and the quality scores were computed by aggregating the results of all pairs. We used hyper-parameters same as the ones suggested in the original paper~\cite{chu2015video}. 

{\bf Collaborative Video Summarization (CVS)} \cite{Panda_2017_CVPR}: CVS is the method that computes a representation of a video clip based on sparse modeling, similar to SMRS. The main difference is that CVS aims to extract a summary that is representative of other videos belonging to the same group as well as the video. We selected the clips whose $l_2$ norm of representation was the largest. The decision of hyper-parameters follows the original paper~\cite{Panda_2017_CVPR}.

{\bf Weakly Supervised Video Summarization (WSVS)} \cite{Panda_2017_ICCV} :  
Similar to our method, WSVS creates a summary using multiple groups. It computes the importance score by calculating the gradient of the classification network with regard to the input space, and aggregating it over a clip. The techniques for training the classification network such as network structure, learning setting, and data augmentation, followed the original paper~\cite{Panda_2017_ICCV}. For a fair comparison, we leveraged the same network as the one we used as well as the one proposed in the original paper pre-trained on split-1 of the UCF101~\cite{soomro2012ucf101} dataset (denoted as WSVS (large) and WSVS respectively).
Moreover, all clips were used for training, and gradients were calculated for them.

\begin{table*}[h!]
\small
\centering
  \caption{Top-5 mean AP computed from human-created summary and predicted summary for each method. Results are shown for each {\bf target group}. For referring to the abbreviated names of groups, please see the Table~\ref{tab:table1}.}
  \vspace{-0.2cm}
\begin{tabular}{c||cccccccccc|c}
\hline
{} &     RV &     RB &     BS &     DS &     RD &     SR &     CC &     RN &     SC &     RS &   mean \\
\hline
SMRS~\cite{elhamifar2012see}                &  0.318 &  0.371 &  0.338 &  0.314 &  0.283 &  0.317 &  0.294 &  \underline{0.348} &  0.348 &  0.286 &  0.322 \\
CK                      &  0.329 &  0.321 &  0.291 &  0.269 &  0.318 &  0.271 &  0.275 &  0.295 &  0.305 &  0.268 &  0.294 \\
CS                      &  0.318 &  0.330 &  0.309 &  0.317 &  0.278 &  0.293 &  0.302 &  \bf{0.355} &  0.350 &  0.271 &  0.312 \\
MBF~\cite{chu2015video}                     &  \bf{0.387} &  0.332 &  0.345 &  0.316 &  0.319 &  0.324 &  \underline{0.375} &  0.317 &  0.324 &  0.288 &  0.333 \\
CVS~\cite{Panda_2017_CVPR}                   &  0.339 &  0.365 &  \bf{0.388} &  0.334 &  \underline{0.359} &  0.386 &  0.362 &  0.303 &  0.337 &  0.356 &  0.353 \\
WSVS~\cite{Panda_2017_ICCV}                    &  0.333 &  0.339 &  0.310 &  0.331 &  0.272 &  0.335 &  0.336 &  0.303 &  0.329 &  0.330 &  0.322 \\
WSVS (large)~\cite{Panda_2017_ICCV}            &  0.331 &  0.350 &  0.322 &  0.294 &  0.304 &  0.306 &  0.308 &  0.322 &  0.342 &  0.310 &  0.319 \\ \hline
ours                    &  \underline{0.373} &  \bf{0.382} &  \underline{0.367} &  \underline{0.396} &  0.327 &  \underline{0.497} &  \underline{0.374} &  0.340 &  \underline{0.368} &  \underline{0.368} &  \underline{0.379} \\
ours (feature learning) &  \underline{0.372} &  \underline{0.376} &  0.299 &  \bf{0.403} &  \bf{0.373} &  \bf{0.518} &  \bf{0.388} &  0.338 &  \bf{0.408} &  \bf{0.378} &  \bf{0.385} \\
\hline
\end{tabular}
\label{fig:exp1}
\end{table*}

The top-5 MAP are shown in Table~\ref{fig:exp1}. 
First, our method performed better than the other methods, which consider only the representativeness from a single group, in most of the {\bf target groups}, and showed competitive performance in the other. 
It implies that discriminative information is the key to estimating the {\it viewpoint}.
\begin{table}[t!]
\vspace{-0.5cm}
\small 
  \centering
  \caption{User study results for the quality evaluation.}
  \vspace{-0.25cm}
  \label{tab:table-statistics}
   \begin{tabular}{|c||c|c|c|c|} \hline 
    method & MBF~\cite{chu2015video} & CVS~\cite{Panda_2017_CVPR}  & ours\\ \hline 
    score & 1.07 & 1.22 & \bf{1.32} \\ \hline 
   \end{tabular}
     \vspace{0.3cm}
   \label{tb:user}
\end{table}

Secondly, the performance of our methods with feature learning was better than that without it as a whole. We found it works well even though we exploited a large network with enormous parameters and the number of samples was relatively small in many cases, except in a few categories.
When considering ``riding bike on beach (RB)'' or ``boarding on a snow mountain (BS)'', we noticed a drop in the performance.  
Our feature learning algorithm works in a kind of self-supervised manner; It trains the feature extractor to explain the current summary better, and therefore, it is dependent on the initial summary selection. If outliers have a high importance score in that step, no matter whether it is discriminative, the parameter update is likely to be strongly affected by such outliers, which causes a performance drop.

Thirdly, we found the performance of WSVS and WSVS (large) were worse than our method and even than CSV, which uses only one group. We assume the reason is that it failed to train the classification model. This method trains the classification model clip-by-clip by assigning the same label to all video clips. It implicitly assumes all clips can be classified into the same group, which is unrealistic when using generic videos such as ones on the web as stated in Section~\ref{sec:relatedwork}. 
If there are some clips that are difficult or impossible to classify, it suffers from over-fitting caused by attempting to correctly classify them.
In our case, we assume there are multiple possible ways to divide videos into groups given the same set of videos, as stated earlier.
Therefore, parameters cannot be appropriately learned because some clips in videos belonging to different groups can appear to be similar. 
Given that our method considers the discrimination of the generated summary, not all clips, it worked better even when using CNN with large parameters.

\subsection{User study}
Because video summarization is a relatively subjective task, we also evaluated the performance with a user study. We asked crowd-workers to assign the quality score to summaries generated from MBF, CVS, and proposed method.
They chose the score from -2 (bad) to 2 (good), and for each video and concept, 10 workers were assigned. The mean results are shown in Table~\ref{tb:user}. It indicates that the quality of summaries of our method is the best among three methods.

\subsection{Visualizing the reason of group division}
One possible application of our method is visualizing the reason driving group divisions. Given multiple groups of videos, why they are grouped in such way is unknown, our algorithm works to visualize an underlying visual concept that is a criterion of the division. To determine how well our algorithm has the ability of this, we performed a qualitative evaluation using AMT.
We asked crowd-workers to select the topic out of either \textbf{concept1} or \textbf{concept2} for summaries created in the group setting1 and setting2. We evaluated the performance of how well workers can answer questions about a topic correctly.
We set the ground-truth topic as \textbf{concept1} when setting1 was used and \textbf{concept2} for setting2.
We assigned 10 workers for each summary and each setting.
As shown in the Table~\ref{tb:estm}, our method performed better than other methods, which indicates the ability to explain the reason behind grouping.

\section{Conclusion}\label{sec:conclusion}
In this study, we introduced a {\it viewpoint} for video summarization motivated by the claim that multiple optimal summaries should exist for one video. We developed a general video summarization method that aims to estimate underlying {\it viewpoint} by considering video-level {\it similarity} which is assumed to be derived from corresponding {\it viewpoint}.
For the evaluation, we compiled a novel dataset and demonstrated the effectiveness of proposed method by performing the qualitative and quantitative experiments on it.

\begin{table}[t!]
\vspace{-0.5cm}
\small 
  \centering
  \caption{User study results for topic selection task. The accuracy takes the value in the range $[0, 1]$.}
  \vspace{-0.2cm}
  \label{tab:table-statistics}
   \begin{tabular}{|c||c|c|c|c|} \hline 
    method & MBF~\cite{chu2015video} & CVS~\cite{Panda_2017_CVPR}  & ours\\ \hline 
    accuracy & 0.47 & 0.60 & \bf{0.76} \\ \hline 
   \end{tabular}
   \vspace{0.2cm}
   \label{tb:estm}
\end{table}

\section{Acknowledgement}
This work was partially supported by JST CREST Grant Number JPMJCR1403, Japan. This work was also partially supported by the Ministry of Education, Culture, Sports, Science and Technology (MEXT)  as ``Seminal Issue on Post-K Computer.''

{\small
\bibliographystyle{ieee}
\bibliography{egbib}

\begin{thebibliography}{10}\itemsep=-1pt

\bibitem{chen2011formulating}
F.~Chen and C.~De~Vleeschouwer.
\newblock Formulating team-sport video summarization as a resource allocation
  problem.
\newblock {\em TCSVT}, 21(2):193--205, 2011.

\bibitem{chu2015video}
W.-S. Chu, Y.~Song, and A.~Jaimes.
\newblock Video co-summarization: Video summarization by visual co-occurrence.
\newblock In {\em CVPR}, 2015.

\bibitem{doersch2013mid}
C.~Doersch, A.~Gupta, and A.~A. Efros.
\newblock Mid-level visual element discovery as discriminative mode seeking.
\newblock In {\em NIPS}, 2013.

\bibitem{doersch2012makes}
C.~Doersch, S.~Singh, A.~Gupta, J.~Sivic, and A.~Efros.
\newblock What makes paris look like paris?
\newblock {\em ACM Transactions on Graphics}, 31(4), 2012.

\bibitem{doersch2015makes}
C.~Doersch, S.~Singh, A.~Gupta, J.~Sivic, and A.~A. Efros.
\newblock What makes paris look like paris?
\newblock {\em Communications of the ACM}, 58(12), 2015.

\bibitem{elhamifar2017online}
E.~Elhamifar and M.~C. D.~P. Kaluza.
\newblock Online summarization via submodular and convex optimization.
\newblock In {\em IEEE Conference on Computer Vision and Pattern Recognition},
  2017.

\bibitem{elhamifar2012see}
E.~Elhamifar, G.~Sapiro, and R.~Vidal.
\newblock See all by looking at a few: Sparse modeling for finding
  representative objects.
\newblock In {\em CVPR}, 2012.

\bibitem{fleischman2007temporal}
M.~Fleischman, B.~Roy, and D.~Roy.
\newblock Temporal feature induction for baseball highlight classification.
\newblock In {\em ACMMM}, 2007.

\bibitem{gong2014diverse}
B.~Gong, W.-L. Chao, K.~Grauman, and F.~Sha.
\newblock Diverse sequential subset selection for supervised video
  summarization.
\newblock In {\em NIPS}, 2014.

\bibitem{gb08}
M.~Grant and S.~Boyd.
\newblock Graph implementations for nonsmooth convex programs.
\newblock In {\em Recent Advances in Learning and Control}. 2008.
\newblock \url{http://stanford.edu/~boyd/graph_dcp.html}.

\bibitem{cvx}
M.~Grant and S.~Boyd.
\newblock {CVX}: Matlab software for disciplined convex programming, version
  2.1.
\newblock \url{http://cvxr.com/cvx}, Mar. 2014.

\bibitem{GygliECCV14}
M.~Gygli, H.~Grabner, H.~Riemenschneider, and L.~Van~Gool.
\newblock Creating summaries from user videos.
\newblock In {\em ECCV}, 2014.

\bibitem{gygli2015video}
M.~Gygli, H.~Grabner, and L.~Van~Gool.
\newblock Video summarization by learning submodular mixtures of objectives.
\newblock In {\em CVPR}, 2015.

\bibitem{hong2009event}
R.~Hong, J.~Tang, H.-K. Tan, S.~Yan, C.~Ngo, and T.-S. Chua.
\newblock Event driven summarization for web videos.
\newblock In {\em SIGMM workshop}, 2009.

\bibitem{jain2013representing}
A.~Jain, A.~Gupta, M.~Rodriguez, and L.~S. Davis.
\newblock Representing videos using mid-level discriminative patches.
\newblock In {\em CVPR}, 2013.

\bibitem{karpathy2014large}
A.~Karpathy, G.~Toderici, S.~Shetty, T.~Leung, R.~Sukthankar, and L.~Fei-Fei.
\newblock Large-scale video classification with convolutional neural networks.
\newblock In {\em CVPR}, 2014.

\bibitem{khosla2013large}
A.~Khosla, R.~Hamid, C.-J. Lin, and N.~Sundaresan.
\newblock Large-scale video summarization using web-image priors.
\newblock In {\em CVPR}, 2013.

\bibitem{kim2014joint}
G.~Kim, L.~Sigal, and E.~P. Xing.
\newblock Joint summarization of large-scale collections of web images and
  videos for storyline reconstruction.
\newblock In {\em CVPR}, 2014.

\bibitem{kulesza2012determinantal}
A.~Kulesza, B.~Taskar, et~al.
\newblock Determinantal point processes for machine learning.
\newblock {\em Foundations and Trends{\textregistered} in Machine Learning},
  5(2--3):123--286, 2012.

\bibitem{lanckriet2009convergence}
G.~R. Lanckriet and B.~K. Sriperumbudur.
\newblock On the convergence of the concave-convex procedure.
\newblock In {\em NIPS}, 2009.

\bibitem{lee2012discovering}
Y.~J. Lee, J.~Ghosh, and K.~Grauman.
\newblock Discovering important people and objects for egocentric video
  summarization.
\newblock In {\em CVPR}, 2012.

\bibitem{li2015mid}
Y.~Li, L.~Liu, C.~Shen, and A.~van~den Hengel.
\newblock Mid-level deep pattern mining.
\newblock In {\em CVPR}, 2015.

\bibitem{liu2010hierarchical}
D.~Liu, G.~Hua, and T.~Chen.
\newblock A hierarchical visual model for video object summarization.
\newblock {\em TPAMI}, 32(12):2178--2190, 2010.

\bibitem{liu2002optimization}
T.~Liu and J.~R. Kender.
\newblock Optimization algorithms for the selection of key frame sequences of
  variable length.
\newblock In {\em ECCV}, 2002.

\bibitem{lu2013story}
Z.~Lu and K.~Grauman.
\newblock Story-driven summarization for egocentric video.
\newblock In {\em CVPR}, 2013.

\bibitem{ma2002user}
Y.-F. Ma, L.~Lu, H.-J. Zhang, and M.~Li.
\newblock A user attention model for video summarization.
\newblock In {\em ACMMM}, 2002.

\bibitem{mahasseni2017unsupervised}
B.~Mahasseni, M.~Lam, and S.~Todorovic.
\newblock Unsupervised video summarization with adversarial lstm networks.
\newblock In {\em CVPR}, 2017.

\bibitem{ngo2003automatic}
C.-W. Ngo, Y.-F. Ma, and H.-J. Zhang.
\newblock Automatic video summarization by graph modeling.
\newblock In {\em ICCV}, 2003.

\bibitem{Panda_2017_ICCV}
R.~Panda, A.~Das, Z.~Wu, J.~Ernst, and A.~K. Roy-Chowdhury.
\newblock Weakly supervised summarization of web videos.
\newblock In {\em ICCV}, 2017.

\bibitem{Panda_2017_CVPR}
R.~Panda and A.~K. Roy-Chowdhury.
\newblock Collaborative summarization of topic-related videos.
\newblock In {\em CVPR}, 2017.

\bibitem{plummer2017enhancing}
B.~A. Plummer, M.~Brown, and S.~Lazebnik.
\newblock Enhancing video summarization via vision-language embedding.
\newblock In {\em Computer Vision and Pattern Recognition}, 2017.

\bibitem{potapov2014category}
D.~Potapov, M.~Douze, Z.~Harchaoui, and C.~Schmid.
\newblock Category-specific video summarization.
\newblock In {\em ECCV}, 2014.

\bibitem{sharghi2016query}
A.~Sharghi, B.~Gong, and M.~Shah.
\newblock Query-focused extractive video summarization.
\newblock In {\em ECCV}, 2016.

\bibitem{sharghi2017query}
A.~Sharghi, J.~S. Laurel, and B.~Gong.
\newblock Query-focused video summarization: Dataset, evaluation, and a memory
  network based approach.
\newblock In {\em 2017 IEEE Conference on Computer Vision and Pattern
  Recognition (CVPR)}, pages 2127--2136. IEEE, 2017.

\bibitem{singh2012unsupervised}
S.~Singh, A.~Gupta, and A.~A. Efros.
\newblock Unsupervised discovery of mid-level discriminative patches, 2012.

\bibitem{song2015tvsum}
Y.~Song, J.~Vallmitjana, A.~Stent, and A.~Jaimes.
\newblock Tvsum: Summarizing web videos using titles.
\newblock In {\em CVPR}, 2015.

\bibitem{soomro2012ucf101}
K.~Soomro, A.~R. Zamir, and M.~Shah.
\newblock Ucf101: A dataset of 101 human actions classes from videos in the
  wild.
\newblock {\em arXiv preprint arXiv:1212.0402}, 2012.

\bibitem{sun2014ranking}
M.~Sun, A.~Farhadi, and S.~Seitz.
\newblock Ranking domain-specific highlights by analyzing edited videos.
\newblock In {\em ECCV}, 2014.

\bibitem{tran2015learning}
D.~Tran, L.~Bourdev, R.~Fergus, L.~Torresani, and M.~Paluri.
\newblock Learning spatiotemporal features with 3d convolutional networks.
\newblock In {\em ICCV}, 2015.

\bibitem{yuille2002concave}
A.~L. Yuille and A.~Rangarajan.
\newblock The concave-convex procedure (cccp).
\newblock In {\em NIPS}, 2002.

\bibitem{yuille2003concave}
A.~L. Yuille and A.~Rangarajan.
\newblock The concave-convex procedure.
\newblock {\em Neural computation}, 15(4):915--936, 2003.

\bibitem{zhang2016summary}
K.~Zhang, W.-L. Chao, F.~Sha, and K.~Grauman.
\newblock Summary transfer: Exemplar-based subset selection for video
  summarization.
\newblock In {\em CVPR}, 2016.

\bibitem{zhu2007trajectory}
G.~Zhu, Q.~Huang, C.~Xu, Y.~Rui, S.~Jiang, W.~Gao, and H.~Yao.
\newblock Trajectory based event tactics analysis in broadcast sports video.
\newblock In {\em ACMMM}, 2007.

\end{thebibliography}
}

\end{document}